# Morphology Generation for Statistical Machine Translation

**Sreelekha S, Pushpak Bhattacharyya**
Indian Institute of Technology (IIT) Bombay, India
{sreelekha, pb}@cse.iitb.ac.in

**Abstract**
When translating into morphologically rich languages, Statistical MT approaches face the problem of data sparsity. The severity of the sparseness problem will be high when the corpus size of morphologically richer language is less. Even though we can use factored models to correctly generate morphological forms of words, the problem of data sparseness limits their performance. In this paper, we describe a simple and effective solution which is based on enriching the input corpora with various morphological forms of words. We use this method with the phrase-based and factor-based experiments on two morphologically rich languages: Hindi and Marathi when translating from English. We evaluate the performance of our experiments both in terms automatic evaluation and subjective evaluation such as adequacy and fluency. We observe that the morphology injection method helps in improving the quality of translation. We further analyze that the morph injection method helps in handling the data sparseness problem to a great level.

**Keywords:** Morphology, Machine Translation, Morphology Injection

## 1. Introduction

Factored models (Koehn and Hoang, 2007; Tamchyna and Bojar, 2013) treat each word in the corpus as vector of tokens. Each token can be any linguistic information about the word which leads to its inflection on the target side. Hence, factored models are preferred over phrase based models (Koehn, Och and Marcu, 2003) when translating from morphologically poor language to morphologically richer language (Avramidis and Koehm, 2008; Chahuneau et al., 2013). Factored models translate using *Translation* and *Generation* mapping steps. If a particular factor combination in these mapping steps has no evidence in the training corpus, then it leads to the problem of data sparseness. Hence, though factored models give more accurate morphological translations, but they may also generate more unknowns compared to other unfactored models. Hindi is morphologically complex comparing to English; while Marathi is more complex with its suffix agglutination properties. However Marathi and Hindi have some similarities except in Marathi there is agglutination of suffixes. To understand the severity of the sparseness problem, we consider an example of verb morphology in Marathi. In Marathi, a regular root verb generates over 80 forms and over 53 irregular verb forms. Each verb can have 2268 (4*3*3*2*3*3*7*6) inflected forms of it. Marathi vocabulary has around 40503 root verbs. Hence, in total 91,860,804(2268*40503) verb forms. It is very likely that parallel Marathi corpus cannot have all inflected forms of each verb. Moreover, if the corpus size of Marathi language is less, then the severity of the sparseness problem will be high.

Thus, even though we can use factored models to correctly generate morphological forms of words, the problem of data sparseness limits their performance. In this paper, we propose a simple and effective solution which is based on enriching the input corpora with various morphological forms of words. Application of the technique to English-Hindi and Marathi case-study shows that the technique really improves the translation quality and handles the problem of sparseness effectively. The rest of the paper is organized as follows: We start by studying the basics of factored translation models. We discuss the sparseness problem in Section 2. Then we describe the Morphology Generation and the factored model for handling morphology, a case study of handling morphology for English to Hindi translation in Section 3. We describe our experiments and evaluations performed in Section 4. Section 5 draws conclusion and points to future work.

## 2. Sparseness in factored translation models

While factored models allow incorporation of linguistic annotations, it also leads to the problem of data sparseness. The sparseness is introduced in two ways:

- **Sparseness in Translation:** When a particular combination of factors does not exist in the source side training corpus. For example, let the factored model have single translation step: X|Y→P|Q. Suppose the training data has evidence for only $x_i|y_j \rightarrow p_k|q_l$ mapping. The factored model learnt from this data can not translate $x_u|y_v$, for all u≠i and v≠j. The factored model generates UNKNOWN as output in these cases. Note that, if we train simple phrase based model on only the surface form of words, we will at least get some output, which may not be correctly inflected, but still will be able to convey the meaning.
- **Sparseness in Generation:** When a particular combination of factors does not exist in the target side training corpus. For example, let the factored model have single generation step: $P\ |Q \rightarrow R$. Suppose the tar- get side training data has an evidence of only $p_a/q_b \rightarrow r_c$. The factored model learnt from this data can not generate from $p_u/q_v$ for all *u≠a* and *v≠b*. Again the factored model generates *UNKNOWN* as output. Thus, due to sparseness, we cannot make the best use of factored models. In fact, they fare worse than the phrase-based models, especially, when a particular factor combination is absent in the training data.

A simple and effective solution to the sparseness problem is to have all factor combinations present in the training data. For the factored model described in Section 3.1, in order to remove data sparseness in the translation step, we need to have all *Source root|{S}* → *Target root|suffix* pairs present in the training data. Also, to remove data sparseness in the generation step, we need to have all

*Target root|suffix → Target surface word* pairs present in the training data. In Section 3, we use a solution to this problem in the context of English to Hindi translation.

## 3. Morphology Generation

Hindi is a morphologically richer language compared to English. Hindi shows morphological inflections on nouns and verbs. In this section, we study the problem of handling noun and verb morphology when translating from English to Hindi using factored models. We also discuss the solution to the sparseness problem.

### 3.1 Noun morphology

In this section, we discuss the factored model for handling Hindi noun morphology and the data sparseness solution in the context of same.

#### 3.1.1 Factored model setup

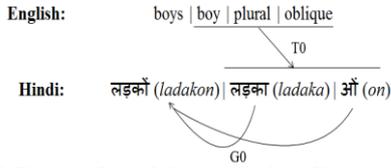

**Figure1: Factored model setup to handle nominal inflections**

Noun inflections in Hindi are affected by the number and case of the noun only (Singh et al., 2010). So, in this case, the set S, consists of number and case. Number can be *singular* or *plural* and case can be *direct* or *oblique*. Example of factors and mapping steps are shown in Figure 1. The generation of the number and case factors is discussed in Section 4.

#### 3.1.2 Building word-form dictionary

In the case of factored model described in Section 3.1.1:

- To solve the sparseness in translation step, we need to have all *English root|number|case → Hindi root noun|suffix* pairs present in the training data.
- To solve the sparseness in generation step, we need to have all *Hindi root noun|suffix → Hindi surface word* pairs present in the training data.

In other words, we need to get a set of suffixes and their corresponding number-case values for each noun pair.

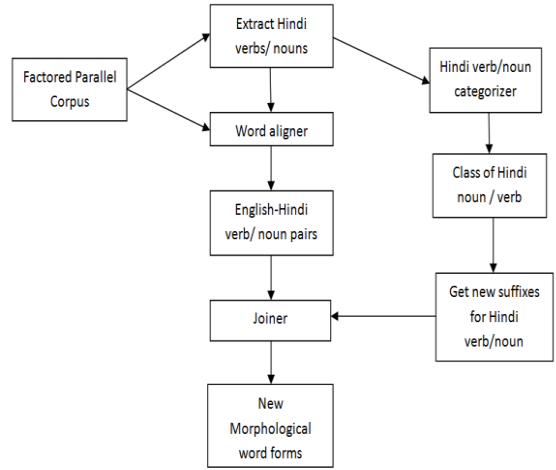

**Figure 2: Pipeline to generate new morphological forms for noun/verbs pair**

Using these suffixes and the Hindi root word, we need to generate Hindi surface words to remove sparseness in the generation step. We need to generate four pairs for each noun present in the training data, i.e., (sg-dir, sg-obl, pl-dir, pl-obl) and get their corresponding Hindi inflections. In the following section, we discuss how to generate these morphological forms.

**Generating new morphological forms:** Figure 2 shows a pipeline to generate new morphological forms for an English-Hindi noun pair. To generate different morphological forms, we need to know the suffix of a noun in Hindi for the corresponding number and case combination. We use the classification table shown in Table 1 for the same. Nouns are classified into five different classes, namely A, B, C, D, and E according to their inflectional behavior with respect to case and number (Singh et al., 2010). All nouns in the same class show the same inflectional behavior. To predict the class of a Hindi noun, we develop a classifier which uses gender and the ending characters of the nouns as features (Singh et al., 2010). We get four different suffixes and corresponding number-case combinations using the class of Hindi noun and classification shown in Table 1. For example, if we know that the noun *form* कुत्ता(*kuttaa*) belongs to class D, then we can get four different suffixes for कुत्ता(*kuttaa*) as shown in Table 2.

|  | Class A | Class B | Class C | Class D | Class E |
|---|---|---|---|---|---|
| **Sg-dir** | *null* | *null* | *null* | *null* | *null* |
| **Sg-obl** | *null* | *null* | *null* | *-e* | *null* |
| **Pl-dir** | *null* | *-yā* | *-ē* | *-e* | *null* |
| **Pl-obl** | *null* | *-yō* | *-ō* | *-ō* | *-yō/-ō* |
| **Examples** | भूख, क्रोध, प्यार | लड़की, शक्ति, नदी | रात, माला, बहू | लड़का, धागा, भांजा | आलू, साधू, माली |

**Table 1: Inflection-based classification of Hindi nouns**

| English root/Number/Case | Hindi surface/Root/Suffix |
|---|---|
| dog/singular/direct | कुत्ता (kuttaa) / कुत्ता (kuttaa)/null |
| dog/singular/oblique | कुत्ते (kutte) / कुत्ता (kuttaa)/e (e) |
| dog/plural/direct | कुत्ते (kutte) / कुत्ता (kuttaa)/e (e) |
| dog/plural/oblique | कुत्तों (kutton) / कुत्ता (kuttaa)/a (on) |

**Table 2: Morphological suffixes for dog-कुत्ता(kuttaa) noun pair**

| English root/Number/Case | Hindi root/Suffix |
|---|---|
| dog/singular/direct | कुत्ता (kuttaa) /null |
| dog/singular/oblique | कुत्ता (kuttaa) / ए (e) |
| dog/plural/direct | कुत्ता (kuttaa) / ए (e) |
| dog/plural/oblique | कुत्ता (kuttaa) / ओं (on) |

**Table 3: New Morphological forms dog-कुत्ता(kuttaa)noun pair**

**Generating surface word:**

Next we generate Hindi surface word from Hindi root noun and suffix using a rule-based joiner (reverse morphological) tool. The rules of the joiner use the ending of the root noun and the class to which the suffix belongs as features. Thus, we get four different morphological forms of the noun entities present in the training data. We augment the original training data with these newly generated morphological forms. Table 3 shows four morphological forms of *dog*-कुत्ता *(ladakaa)* noun pair. The joiner solves the sparseness in generation step.

## 3.2 Verb morphology

In this section, we discuss the factored model for handling Hindi verb morphology and the data sparseness solution in the context of same.

### 3.2.1 Factored model setup

Verb inflections in Hindi are affected by gender, number, person, tense, aspect, modality, etc. (Singh and Sarma, 2011). As it is difficult to extract gender from English verbs, we do not use it as a factor on English side. We just replicate English verbs for each gender inflection on Hindi side. Hence, set S, as in Section 3.1, consists of number, person, tense, aspect and modality. Example of factors and mapping steps are shown in Figure 3. The generation of the factors is discussed in Section 5.4.

### 3.2.2 Building word-form dictionary

Thus, in the case of factored model described in Section 5.2.1:

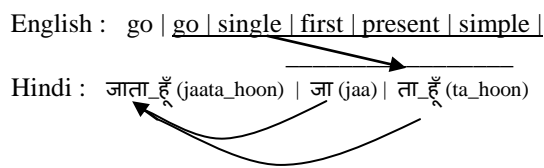

English :   go | go | single | first | present | simple |

Hindi :   जाता_हूँ (jaata_hoon) | जा (jaa) | ता_हूँ (ta_hoon)

**Figure 3: Factored model setup to handle nominal inflections**

- To solve the sparseness in translation step, we need to have all *English root/numer/person/tense/aspect/modality →Hindi root verb|suffix* pairs present in the training data.
- To solve the sparseness in generation step, we need to have all *Hindi root verb|suffix → Hindi surface word* pairs present in the training data.

In other words, we need to get a set of suffixes and their corresponding number-person-tense-aspect-modality values, for each noun pair. Using these suffixes and the Hindi root word, we need to generate Hindi surface words to remove sparseness in the generation step. In the following section, we discuss how to generate these morphological forms.

**Generating new morphological forms:** Figure 2 shows a pipeline to generate new mor- phological forms for an English-Hindi verb pair. No pre-classification of verbs is required, as these suffixes apply to all verbs.

**Generating surface word:** Next we generate Hindi surface word from Hindi root verb and suffix using a rule-based joiner (reverse morphological) tool. The rules of the joiner use only the ending of the root verb as features. Thus, we get different morphological forms of the verb entities present in the training data. We augment the original training data with these newly generated morphological forms. The joiner solves the sparseness in generation step.

## 3.3 Noun and verb morphology

Finally, we create a new factored model which combines factors on both nouns and verbs, as shown in Figure 4. We build word-form dictionaries separately as discussed in Section 3.1.2 and Section 3.2.2. Then, we augment training data with both dictionaries. Note that, factor normalization4 on each word is required before this step to maintain same number of factors. We also create a word-form dictionary for phrase-based model. We follow the same procedure as described above, but we remove all factors from source and target words except the surface form.

## 4. Experiments and Evaluation

We performed experiments on ILCI (Indian Languages Corpora Initiative) En-Hi and En-Mr data set. Domain of the corpus is health and tourism. We used 44,586 sentence pairs for training and 2,974 sentence pairs for testing. Word-form dictionary was created using the Hindi and Marathi word lexicon. It consisted of 182,544 noun forms and 310,392 verb forms of Hindi and 44,762 noun forms and 106,570 verb forms of Marathi. *Moses*[1] toolkit was used for training and decoding. Language model was trained on the target side corpus with *IRSTLM*[2]. For our experiments, we compared the translation output of the following systems: (1) Phrase-based (unfactored) model (**Phr**); (2) Basic factored model for solving noun and verb morphology (**Fact**); (3) Phrase-based model trained on the corpus used for *Phr* augmented with the word form dictionary for solving noun and verb morphology (**Phr'**); (4)Factored model trained on the corpus used for *Fact* augmented with the word form dictionary for solving noun and verb morphology (**Fact'**). With the help of syntactic and morphological tools, we extract the number and case of the English nouns and number, person, tense, aspect and modality of the English verbs as follows:

**Noun factors:**
- **Number factor:** We use *Stanford POS tagger*[3] to identify the English noun entities (Toutanova et al., 2003). The POS tagger itself differentiates between singular and plural nouns by using different tags.
- **Case factor:** It is difficult to find the direct/oblique case of the nouns as English nouns do not contain this information. Hence, to get the case information, we need to find out features of an English sentence that correspond to direct/oblique case of the parallel nouns in Hindi sentence. We use object of preposition, subject, direct object, tense as our features. These features are extracted using semantic relations provided by Stanfords typed dependencies (De Marneffe et al., 2008).

**Verb factors:**
- **Number factor:** Using typed dependencies we extract subject of the sentence and get number of the subject as we get it for a noun.
- **Person factor:** We do lookup into simple list of pronouns to find the person of the subject.

**Tense, Aspect and Modality factor:** We use POS tag of verbs to extract tense, aspect and modality of the sentence.

---
[1] http://www.statmt.org/moses/
[2] https://hlt.fbk.eu/technologies/irstlm-irst-languagemodelling-toolkit
[3] http://nlp.stanford.edu/software/tagger.shtml

| Morph. problem | Model | BLEU En-Hi | BLEU En-Mr | # OOV En-Hi | # OOV En-Mr | % OOV reduction En-Hi | % OOV reduction En-Mr | Adequacy En-Hi | Adequacy En-Mr | Fluency En-Hi | Fluency En-Mr |
|---|---|---|---|---|---|---|---|---|---|---|---|
| Noun | *Fact* | 25.30 | 12.84 | 2,130 | 1,499 | | | 25.62 | 16.52 | 25.65 | 17.20 |
| | *Fact'* | 28.41 | 17.86 | 1,839 | 1,302 | **19.33** | **15.08** | 30.73 | 23.58 | 35.66 | 26.25 |
| Verb | *Fact* | 26.23 | 13.02 | 1,241 | 1,872 | | | 26.85 | 18.67 | 27.86 | 19.26 |
| | *Fact'* | 29.16 | 19.02 | 1,010 | 1,649 | **20.11** | **17.58** | 35.91 | 26.74 | 39.91 | 29.31 |
| Noun & Verb | *Fact* | 22.93 | 10.55 | 2,293 | 3,237 | | | 20.89 | 13.69 | 24.92 | 16.28 |
| | *Fact'* | 24.03 | 12.01 | 1,967 | 2,422 | 1**8.87** | **10.98** | 24.19 | 18.79 | 28.06 | 22.36 |
| Noun & Verb | *Phr* | 24.87 | 13.40 | 913 | 875 | **12.38** | **10.06** | 23.07 | 17.70 | 25.90 | 18.24 |
| | *Phr'* | 27.89 | 16.41 | 853 | 828 | | | 27.15 | 21.72 | 31.92 | 25.25 |

Table 4: Automatic and Subjective evaluation of the translation systems

### 4.1 Automatic Evaluation

The translation systems were evaluated by BLEU score (Papineni et al., 2002). We counted the number of OOV words in the translation outputs, since the reduction in number of unknowns in the translation output indicates better handling of data sparsity. Table 4 shows the evaluation scores and numbers. From the evaluation scores, it is very evident that *Fact'/Phr'* outperforms *Fact/Phr* while solving any morphology problem in both Hindi and Marathi. But, improvements in *En-Mr* systems are low. This is due to the small size of word-form dictionaries that are used for injection. % reduction in OOV shows that, morphology injection is more effective with factored models than with the phrase-based model. Also, improvements shown by BLEU are less compared to % reduction in OOV.

### 4.2 Subjective Evaluation

As BLEU evaluation with single reference is not a true measure of evaluating our method, we also performed human evaluation. We found out that Fact'/Phr' systems really have better outputs compared to Fact/Phr systems, in terms of both, adequacy and fluency. We have randomly chosen 150 translation outputs from each system were manual evaluation to get the adequacy and fluency scores. The scores were given on the scale of 1 to 5 going from worst to best, respectively (Sreelekha, et.al., 2013). Table 4 shows average scores for each system. We observe upto 9% improvement in adequacy and upto 11% improvement in fluency.

### 5. Conclusion

SMT approaches suffer due to data sparsity when translating into morphologically rich languages. We use morphology injection method to solve this problem by enriching the training data with the missing morphological forms of words. We verify this method with two morphologically rich languages Marathi and Hindi when translating form English. We analyze that morphology injection performs very well and improves the translation quality. We observe huge reduction in number of OOVs and improvement in adequacy and fluency of the translation outputs. This method is more effective when used with factored models than the phrase-based models. The morphology generation process may be difficult for target languages which are morphologically too complex even though the approach of solving data sparsity seems simple. A possible future work is to generalize the approach of morphology generation and verify the effectiveness of morphology injection on morphologically complex languages.

### Acknowledgements

This work is funded by Department of Science and Technology, Govt. of India under Women Scientist Scheme- WOS-A with the project code- SR/WOS-A/ET-1075/2014.